\documentclass[letterpaper]{article}
\usepackage{aaai20}
\usepackage{times}
\usepackage{helvet}
\usepackage{courier}
\usepackage[hyphens]{url}
\usepackage{graphicx}
\usepackage{booktabs}
\usepackage{subcaption}
\def\UrlFont{\rm}
\usepackage{graphicx}
\title{Minibatch Processing in Spiking Neural Networks}
\author{Daniel J. Saunders,\textsuperscript{\rm 1, 2} Cooper Sigrist,\textsuperscript{\rm 1} Kenneth Chaney,\textsuperscript{\rm 3} Robert Kozma,\textsuperscript{\rm 1, 4} Hava T. Siegelmann\textsuperscript{\rm 1}\\
\textsuperscript{\rm 1}BINDS Lab, University of Massachusetts Amherst, Amherst, MA, USA\\
\textsuperscript{\rm 2}Fomoro AI, San Francisco, CA, USA\\
\textsuperscript{\rm 3}GRASP, University of Pennsylvania, Philadelphia, PA, USA\\
\textsuperscript{\rm 4}CLION, University of Memphis, Memphis, TN, USA\\
danjsaund@gmail.com, csigrist@umass.edu, chaneyk@seas.upenn.edu, rkozma@cs.umass.edu, hava@cs.umass.edu\\
}

\begin{document}

\maketitle

\begin{abstract}

Spiking neural networks (SNNs) are a promising candidate for  biologically-inspired and energy efficient computation. However, their simulation is notoriously time consuming, and may be seen as a bottleneck in developing competitive training methods with potential deployment on neuromorphic hardware platforms. To address this issue, we provide an implementation of mini-batch processing applied to clock-based SNN simulation, leading to drastically increased data throughput. To our knowledge, this is the first general-purpose implementation of mini-batch processing in a spiking neural networks simulator, which works with arbitrary neuron and synapse models. We demonstrate nearly constant-time scaling with batch size on a simulation setup (up to GPU memory limits), and showcase the effectiveness of large batch sizes in two SNN application domains, resulting in $\approx$880X and $\approx$24X reductions in wall-clock time respectively. Different parameter reduction techniques are shown to produce different learning outcomes in a simulation of networks trained with spike-timing-dependent plasticity. Machine learning practitioners and biological modelers alike may benefit from the drastically reduced simulation time and increased iteration speed this method enables. Code to reproduce the benchmarks and experimental findings in this paper can be found at \texttt{https://github.com/djsaunde/snn-minibatch}.

\end{abstract}

\section{Introduction}
\label{sec:introduction}

Research into training SNNs for machine learning (ML) tasks has rapidly accelerated in recent years \cite{Pfeiffer2018,Tavanaei2019}. This is due in part to their impressive computational power \cite{Maass1996}, their natural applicability to computation over spatio-temporal signals \cite{Yujie2018}, their biological plausibility -- and, therefore, possibilities for synergy with neuroscience \cite{Marblestone2016} -- along with the promise of low energy consumption and rapid processing time once implemented in neuromorphic hardware \cite{Jeong2018}. Software for the efficient training of these networks is, however, largely undeveloped relative to libraries for the training and deployment of artificial neural networks (ANNs). In particular, existing solutions do not support the independent, parallel processing of data through a single network structure.

Due in part to a general lack of mature software infrastructure, researchers have been hesitant to adopt SNNs for cutting-edge ML experimentation. As a result, the development of training algorithms for SNNs has been slow relative to the proliferation of research on ANNs. With large datasets and the complex neural network models needed to process them, advances in software and hardware technology for ANNs has been critical to enabling their practical training and application. In order to bring SNNs to the technological forefront, similar advances are needed. In this paper, we take another step towards the practical use of SNNs with a general-purpose implementation of GPU-enabled minibatch processing. We argue that SNNs may enjoy similarly widespread applicability once they are made simpler to build, simulate, and train \cite{Wu2019}.

Neurons in spiking neural networks \cite{Maass1997} are set apart from those of ANNs in part by their maintenance of simulation state variables over time, e.g., voltages or refractoriness. That is, neurons in SNNs are \textit{stateful}, whereas those typically used in ANNs are \textit{stateless}, with the notable exception of recurrent neural networks (RNNs), which maintain hidden state over time. Indeed, SNNs can be seen as a special case of RNNs, wherein recurrent processing is carried out by dynamic state variables rather than explicit recurrent connections (although, such recurrent connections may also be used in SNNs) \cite{Neftci2019}. Statelessness implies that multiple inputs (i.e., a \textit{minibatch} \cite{Goyal2017}) can be input in parallel to an ANN and processed independently without any additional memory overhead. However, in SNNs and RNNs, where inputs are processed for a length of time and neurons' state variables often depend on their values in the previous time step, there is no choice but to maintain these variables in memory.

Minibatch processing, both at training and inference time, has a number of useful properties:

\begin{itemize}
    \item \textit{Reduced simulation wall-clock time}: Running multiple simulations in parallel enables processing more data per unit wall-clock time than running one at a time. Using a GPU with enough memory, the amount of data processed per unit wall-clock time can be expected to increase approximately linearly with minibatch size.
    \item \textit{Reduced variance in parameter updates}: Computing parameter updates over minibatches results in less noisy updates than those computed from single examples. This reduces the effect that outliers have on parameter updates. On the other hand, for optimization purposes, the noise resulting from small minibatch sizes may help move a model out of local minima \cite{Bottou2010}.
    \item \textit{Improved generalization}: There is good reason to believe that, in the small-minibatch regime, stochastic gradient descent (SGD) improves the generalization performance of ANNs \cite{Poggio2018}. It may be described intuitively as a ``bagging'' procedure \cite{Breiman1996}, where, by computing parameter updates based on a minibatch of examples, we enforce changes that generalize across across minibatches.
\end{itemize}

While mainly useful for machine learning experimentation, researchers working in biological modeling may also benefit from batched simulation. There is no clear analogue of minibatch processing in neural circuits: although the brain is a highly parallel computing device, neural circuits must process their inputs one at a time. However, the technique is not meant to mimic biological phenomena, but rather to increase computational efficiency. To that end, experimenters can use minibatch processing to simulate multiple, independent trials in order to take trial averages, gather data, or calibrate parameter settings more quickly.

In this paper, we describe a general-purpose implementation of minibatch processing in SNNs. To our knowledge, it is the first of its kind, although minibatching in restricted situations has been discussed in prior work. We supplement this description with a concrete implementation in \texttt{BindsNET}, an open source SNN simulation library \cite{Hazan2018}. Written in the Python programming language on top of the \texttt{PyTorch} deep learning framework \cite{Paszke2017}, \texttt{BindsNET} was built with ease of prototyping and machine learning applications in mind. There is support for processing on CPUs and GPUs, where on the latter, users may see significant speed improvements due to the use of minibatch processing. We also provide experiments which showcase the multifaceted benefits of using minibatch processing in SNNs, and discuss the use different batch-wise parameter reduction techniques in the online learning setting.

GPUs are well-suited to parallelizing many of the mathematical operations needed to simulate spiking networks in \texttt{BindsNET}, e.g., the matrix multiplication used to compute the current incident to post-synaptic neurons based on synapse weights and pre-synaptic spiking activity. In general, operations where a single instruction can be applied to many data can easily be mapped to GPUs and parallelized to a large degree.

\section{Related Work}

\subsection{Minibatch processing}
\label{ssec:related_minibatching}

To our knowledge, ours is the first general-purpose implementation of minibatch processing in SNN simulation. Moreover, it is the first to be implemented in an SNN simulation library, and, importantly, works with all available neuron and synapse types and training methodologies. Perhaps the closest to our work is the implementation in \texttt{NengoDL} \cite{Rasmussen2018}, which does not support minibatch processing with \textit{online learning} rules, i.e., those which compute updates to parameters concurrently with data processing, a key feature of spiking neural networks.

The idea of processing with batches of data for the purpose of training statistical or machine learning models is not a new one \cite{Bertsekas1996}. Indeed, in the original formulation of gradient descent, updates to fitted parameters are computed over the entire training dataset. With increasingly large datasets and limitations on memory, this approach is not always feasible, and so computing stochastic updates over randomly sampled batches of data has become standard practice. Indeed, it has even been argued that small batch sizes are desirable in some cases, where it can improve the stability of training and decrease generalization error on the test data \cite{Luschi2018}.

Several prior works have incorporated bespoke implementations of minibatch processing for restricted types of SNNs. \cite{Ferre2018} describe a binary STDP rule that allows for processing minibatches of data, although it only considers the precise timing of neurons' first spikes, and it involves approximating spiking neurons as rectified linear units (ReLUs). \cite{OConner2016} describe an unusual spiking neural network model that allows both positively and negatively signed ``spikes'' and derive approximations to the back-propagation algorithm, claiming ``...in principle it is possible to do minibatch training'', although their experiments involve one-by-one processing of data points. \cite{Lee2018} pre-train a convolutional SNN layer-wise with STDP, and fine-tune the network's weights for a downstream classification task with back-propagation on low-pass filtered spike trains. The networks are trained with minibatch updates, but it is unclear whether they are computed in parallel, or are instead computed serially and later averaged to produce a minibatch update. \cite{Zenke2018} implement a three-factor learning rule for learning precise spatiotemporal spike patterns which is computed over minibatches of data. \cite{Comsa2019} implement minibatched exact back-propagation for training spike times and synapse weights in a network of spiking neurons that emit single spikes.

Other authors have approximated spiking neurons by smoothing their activation function, so as to incorporate them into ANNs to be trained with the back-propagation algorithm \cite{Hunsberger2016,Huh2018}. Here, minibatch processing is obtained for free as a result of the smooth approximations used. However, it is difficult to describe the neurons in these networks as ``spiking'', in the sense that they do not fire all-or-nothing pulses in the event of a voltage threshold crossing.

\subsection{SNN training methodologies}
\label{ssec:related_training}

Due to their power efficiency and event-based operation, research into methods for training SNNs for machine learning tasks has accelerated. Their non-differentiability, due to the all-or-nothing, discontinuous nature of spiking neurons, has made it impossible to train them with the popular back-propagation algorithm. To deal with this, several general training approaches have been developed for SNNs, to all of which minibatch processing is applicable. We review a number of the most well-known approaches:

\begin{itemize}
    \item \textbf{Local learning rules}: Local learning rules \cite{Ralph1992,Zappacosta2018}, such as Hebbian learning \cite{Hebb1949} and spike-timing-dependent plasticity (STDP) \cite{Markram1997,Bi1998}, operate by updating synaptic strengths as a function of pre- and post-synaptic neural activity and possibly a third, global factor such as dopamine or other neuromodulators \cite{Fremaux2016}. In the context of minibatch processing, updates to synapses can be reduced across the minibatch dimension, effectively increasing the speed of learning and possibly decreasing the step-by-step variability of weight changes.
    \item \textbf{Rate-based gradient methods}: In this setting, the temporal aspects of spikes are ignored, and firing rates are considered in lieu of precise spike timing or ordering. Firing rates are often continuous with respect to neuronal inputs, and can therefore be used in back-propagation calculations \cite{Hunsberger2016,OConner2016,Stromatias2017}.
    \item \textbf{Surrogate gradient methods}: These methods provide an approach for overcoming the difficulties associated with the spiking discontinuity by providing an approximating surrogate gradient for the neuron's spiking nonlinearity \cite{Zenke2018,Yujie2018,Shrestha2018,Neftci2019}. Networks are then trained with gradient descent. One such work argues that their derived rule could be used in minibatch updates \cite{Zenke2018}.
    \item \textbf{Differentiable approximations}: Several prior works \cite{Hunsberger2016,Huh2018} have devised differentiable approximations to spiking neuron models and incorporated them into artificial neural networks. These networks may be trained with minibatch updates, as they ignore the temporal dynamics of spiking neurons \cite{Hunsberger2016}, or incorporate them into recurrent ANNs \cite{Huh2018}.
    \item \textbf{ANN to SNN conversion}: A recent thread of research into deploying spiking neural networks on neuromorphic hardware involves the conversion of trained ANNs to SNNs with little or no loss in performance on classification \cite{Diehl2015b,Rueckauer2017,Rueckauer2018,Sengupta2019,Zhang2019} and reinforcement learning \cite{Patel2019} tasks. ANNs are trained with a variant of minibatch gradient descent, but, once converted to SNNs, these works do not apply minibatch processing. See Table \ref{tab:conversion_results} for a comparison of error rates between ANNs and their converted SNN counterparts, demonstrating that, in principle, SNNs may perform just as well on complex classification tasks as ANNs can.
\end{itemize}

\begin{table}
    \centering
    \caption{Comparison of ANN and converted SNN error rates on popular computer vision benchmarks. All results are taken from \cite{Rueckauer2018}, which reports the lowest conversion error rates across all datasets to date.}
    \resizebox{\columnwidth}{!}{
        \begin{tabular}{c|c|c}
            \toprule
            Dataset                      & ANN error & SNN error \\
            \midrule
            MNIST \cite{LeCun2010}       & 0.56\%    & 0.56\%    \\
            CIFAR-10 \cite{Krizhevsky09} & 8.09\%    & 9.15\%    \\
            ImageNet \cite{Deng09}       & 23.88\%   & 25.40\%   \\
            \bottomrule
        \end{tabular}
    }
    \label{tab:conversion_results}
\end{table}

\section{Implementation}
\label{sec:implementation}

Since certain neuron and synapse models in SNNs maintain various stateful quantities during simulation, for a minibatch size $B$, our implementation duplicates these variables $B$ times at the start of a simulation. During simulation, these time-sensitive variables evolve independently across the batch dimension. Quantities that are not stateful are not duplicated, such as rest and reset voltages, fixed thresholds, voltage decay rates, etc. Adaptive parameters such as connection weights (synaptic strengths) and adaptive voltage thresholds are updated during simulation, but there is only one copy of each of these parameters; updates to them are aggregated across the batch dimension via averaging, summation, or possibly many other reductions, which we will later discuss.

\subsection{Dynamic minibatch size}
\label{ssec:dynamic_minibatch}

Adaptive minibatch sizes are supported. Changes in minibatch size may occur when moving from training to inference; e.g., large amounts of training data may be bundled into minibatches to expedite training, whereas at inference time, queries to the trained SNN may occur one at a time as needed. It may also change when the size of a dataset is not evenly divisible by the minibatch size, and so the last batch of examples will be smaller than the rest. Adaptive minibatch size is implemented by checking the batch size of an input against the expected batch size; if it is different, state variables are re-initialized to match it, and simulation proceeds as normal.

\subsection{Episodic vs. continuing simulation}
\label{ssec:episodic_continuing}

Implicit in our discussion thus far is a reliance on episodic, trial-based experimentation. Between trials (processing a minibatch of size B for time T), time-sensitive neuronal state variables must be reset to common values; otherwise, we are not performing $B$ independent simulations with the same initial conditions. This setup is well suited to many machine learning tasks: unsupervised, supervised, and episodic reinforcement learning proceed on a example-by-example or episode-by-episode basis.

However, if the user is comfortable with relaxing the assumption of identical initial conditions, continuing simulations may be used, where input data may change over time without requiring the re-initialization of state variables. This is well-suited for cases where said state variables are relatively transitory, and when their initial conditions don’t have a strong effect on the measured simulation outcomes. For example, after a short simulation time, neuron voltages may change quite rapidly, and it is difficult to guess at their initial values. Continuing simulation may be used for batched continuing reinforcement learning, or for SNN simulations which have no natural notion of ``resetting''.

\subsection{Reduction methods}
\label{ssec:reduction}

It is common practice to average updates to an ANN's parameters over the batch dimension. Every neuron in an ANN participates during the network's forward pass, and averaging the weight updates over the batch dimension results in an unbiased estimate of its derivative with respect to the loss function. The neurons of spiking neural networks, on the other hand, output non-zero values (spikes) relatively sparsely in time, which often trigger parameter updates that have no bearing on a global loss function. Therefore, averaging parameter updates over the batch dimension may result in overly conservative parameter updates and slow learning due to the presence of many zero values in the average, and which can be avoided when not training with gradient descent.

For this reason, our implementation supports arbitrary reduction methods, namely, those that process \texttt{PyTorch} tensors and may be used to reduce the minibatch dimension. Custom reduction methods may be written by users as long as they support this simple API. By default, parameter updates are averaged over the batch dimension. As we will discuss, different applications may benefit from using different reduction methods.

\subsection{Complexity}
\label{ssec:complexity}

Duplicating stateful variables across the batch dimension may quickly consume memory. For per-neuron variables (e.g., membrane voltage), assuming a minibatch size of $B$ and a neuron population of size $N$, $\mathcal{O} (B N)$ memory is required. For per-synapse variables (e.g., synapse conductances), assuming pre- and post-synaptic neuron populations of size $N_\textrm{pre}$ and $N_\textrm{post}$, respectively, $\mathcal{O} (B N_\textrm{pre} N_\textrm{post})$ is needed. Multiple stateful variables per network component may need to be extended across the batch dimension, the number of which generally increases with the complexity of the neuron or synapse model. Users must be wary of setting batch sizes such that the total memory usage is greater than what is available, so as to prevent frequent swapping of tensors in and out of memory or triggering out-of-memory errors. As a result, in comparison with ANNs, minibatch processing in SNNs is fundamentally more memory-intensive due to the use of stateful, time-dependent variables.

It is well-established that GPUs are suited for highly parallel processing due to their large number of cores, which all execute the same instructions simultaneously. For this reason, we expect that the wall-clock time for a given simulation with an SNN of fixed size will remain roughly constant with increasing batch size, up until the point where network variables no longer fit into GPU memory, at which point simulation time will increase as tensors will needed to be swapped in and out of GPU memory. This will be shown empirically in \ref{sec:experiments}.

\section{Experiments}
\label{sec:experiments}

In the following, we describe a few simple experiments aimed at communicating the usefulness of the minibatch processing approach to SNNs simulation. We investigate the scaling of a simple two-layer network to increasing output layer and minibatch sizes. We then show how a simple multi-layer perceptron converted to a near-equivalent SNN can maintain accuracy and classify test data increasingly rapidly with increasing batch size. Finally, SNNs of fixed size are trained in an semi-supervised fashion to classify the MNIST dataset, effective for a wide range of minibatch sizes. Unless otherwise stated, a 1ms simulation time resolution is used.

\subsection{Scaling a Two Layer Network}
\label{ssec:scaling}

We construct a simple two layer network consisting of 100 input neurons with Poisson spike trains with rates randomly sampled in [0Hz, 120Hz] connected to a variable-sized layer of leaky integrate-and-fire (LIF) neurons \cite{Gerstner2002} with synapse weights randomly sampled from $\mathcal{N} (0.1, 0.01)$. Varying the minibatch size, we run the network for 1 second of simulated time in 10 independent trials and report the statistics of the required wall-clock time.

\begin{figure}
    \centering
    \includegraphics[width=\linewidth]{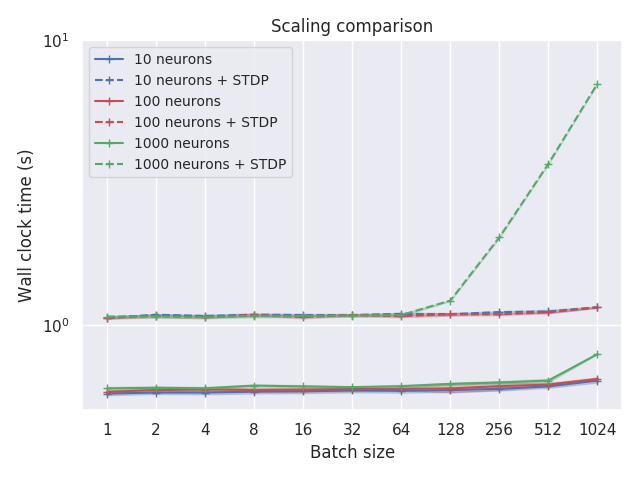}
    \caption{Wall-clock time of a 1s simulation vs. batch size with varying numbers of output neurons, with and without STDP. 10 independent trials are run on a GeForce GTX TITAN X, and their average time $\pm$ one standard deviation is reported. Increase in simulation time for large networks and batch sizes is due to running out of GPU memory (12GB).}
    \label{fig:scaling}
\end{figure}

Figure \ref{fig:scaling} depicts the results for networks with a variable number of output neurons, with or without training the synapse weights with a simple online STDP rule. In all cases, simulation wall-clock time remains roughly constant for small- and medium-sized batch sizes, but begins to grow quickly as the batch size grows large. This is due to running out of GPU memory (12GB) with larger network and minibatch sizes and using STDP. Learning with STDP incurs a higher memory and computational cost, from recording the ``spike traces'' in the pre- and post-synaptic populations required for online STDP, and from computing weight updates and reducing them across the batch dimension.

\subsection{ANN to SNN conversion}

Following the methodology of \cite{Rueckauer2018}, we first train a 3-layer multi-layer perceptron to classify the MNIST data and convert it to an SNN with little loss in performance. The network has hidden layers with sizes of 256 and 128 and ReLU activations. It is converted into an spiking neural network with identical architecture, except that the ReLU non-linearities are approximated by the firing rates of (non-leaky) integrate-and-fire (IF) neurons with \textit{reset by subtraction}. That is, instead of resetting neuron voltages back to a baseline value after a spike (typically zero), the difference between the firing threshold and baseline value is subtracted off the neuron's voltage. This ensures that, if a neuron exceeds its threshold by some amount, that amount is not lost by the resetting mechanism. To derive a classification decision from the network, we sum the inputs to the final layer (with size equal to the number of classes) over the simulation run, and take the label corresponding to the maximizing argument.

\begin{figure*}
    \centering
    \begin{subfigure}[t]{0.475\textwidth}
        \centering
        \includegraphics[width=\textwidth]{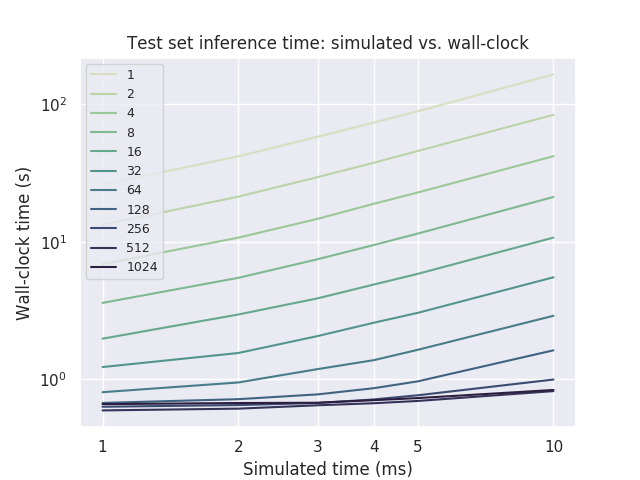}
        \caption{Test set inference time}
        \label{subfig:conversion_inf_time}
    \end{subfigure}
    \begin{subfigure}[t]{0.475\textwidth}
        \centering
        \includegraphics[width=\textwidth]{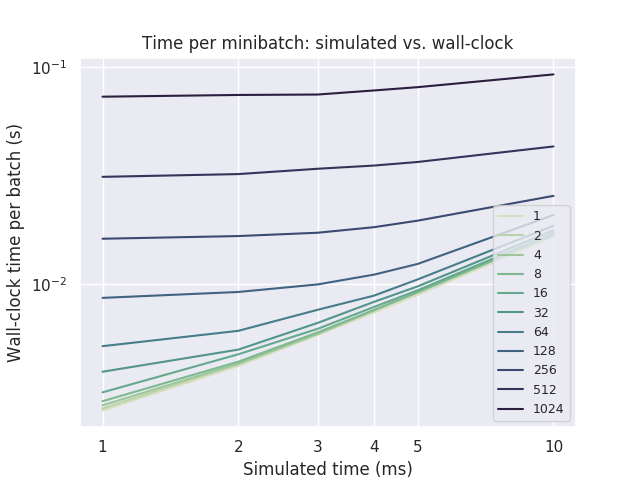}
        \caption{Batch inference time}
        \label{subfig:conversion_batch_time}
    \end{subfigure}
    \caption{(a) Wall-clock time required to classify the test dataset with the converted SNN with various settings of batch size and simulation time. (b) Inference time for a single batch of data for various settings of batch size and simulation time.}
\end{figure*}

Accuracy of the converted SNN compared to the original ANN is given in Table \ref{tab:conversion_accuracy}. The ANN achieves 98.13\% test accuracy, while the SNN with 10ms inference time achieves 97.86\%, a $0.27\%$ reduction. With 3ms of simulation time, the converted SNN already achieves 97.30\% accuracy. Setting the simulation time higher than 10ms does not result in better performance (data not shown). Figure \ref{subfig:conversion_inf_time} plots the wall-clock time required to run inference in the converted SNN on the entire MNIST test dataset (10K images). With batch size 1 (serial processing) and 10ms of simulation time, inference takes over 11 minutes. On the other hand, with batch size 1024, this same procedure takes $\approx$0.75 seconds, a $\approx$880X reduction in wall-clock time. Finally, inference time per minibatch for various settings of batch size and simulation time is plotted in Figure \ref{subfig:conversion_batch_time}. For small batch sizes, each simulation time step takes $\ll$0.01s, while for larger batch sizes, each step requires between 0.01 and 0.1 seconds. A single simulation step with batch size 1 requires just over 1ms of wall clock time, running nearly in ``real time''.

\begin{table}
    \centering
    \caption{Simulation time per example vs. overall accuracy on the MNIST test dataset. The original MLP achieves 98.13\% accuracy.}
    \resizebox{\columnwidth}{!}{
        \begin{tabular}{c|c|c|c|c|c|c}
            \toprule
            Time     & 1ms     & 2ms     & 3ms     & 4ms     & 5ms     & 10ms \\
            Accuracy & 29.37\% & 94.03\% & 97.30\% & 97.62\% & 97.73\% & 97.86\% \\
            \bottomrule
        \end{tabular}
    }
    \label{tab:conversion_accuracy}
\end{table}

\subsection{Unsupervised Learning of MNIST digits}

We implemented a slightly modified, minibatched version of the experimental setup from \cite{Diehl2015a}. The considered SNN consists of an input layer with dimensionality equal to the input data, in this case, the MNIST digits, with shape 28$\times$28. The input data is encoded into Poisson spike trains with firing rates in [0Hz, 128Hz], obtained by dividing the pixel-wise input data by 2. This layer connects all-to-all with STDP-modifiable synapses to a population of $n_\textrm{neurons}$ LIF neurons with adaptive thresholds, which increase by 0.05mV each time a spike is emitted, and are otherwise decaying back to their default value with a time constant of 1000s. This layer is recurrently connected with large, fixed inhibitory synapses, which is used to implement a soft winner-take-all (WTA) circuit: when a neuron in this layer spikes, all other neurons in the layer are inhibited, allowing it to continue spiking unchallenged. Accordingly, we use an online version of STDP (i.e., weight updates are made during simulation) which utilizes only positive weight updates triggered by the firing of the post-synaptic neuron, along with a weight normalization technique such that the sum of weights incident to a post-synaptic neuron remains constant. We implement a simple classification scheme on the output of the network; namely, individual neurons are assigned labels according to the class of data for which they fire most for during training. At test time, spikes are counted per neuron and aggregated into class-wise bins. The bin with the largest count determines the label of the input data.

\begin{figure*}
    \centering
    \begin{subfigure}[t]{0.475\textwidth}
        \centering
        \includegraphics[width=\textwidth]{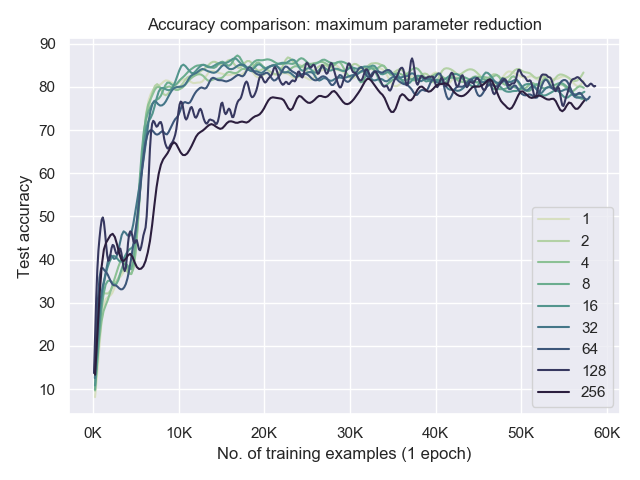}
        \caption{Parameter reduction: maximum}
        \label{subfig:max_training_curves}
    \end{subfigure}
    \begin{subfigure}[t]{0.475\textwidth}
        \centering
        \includegraphics[width=\textwidth]{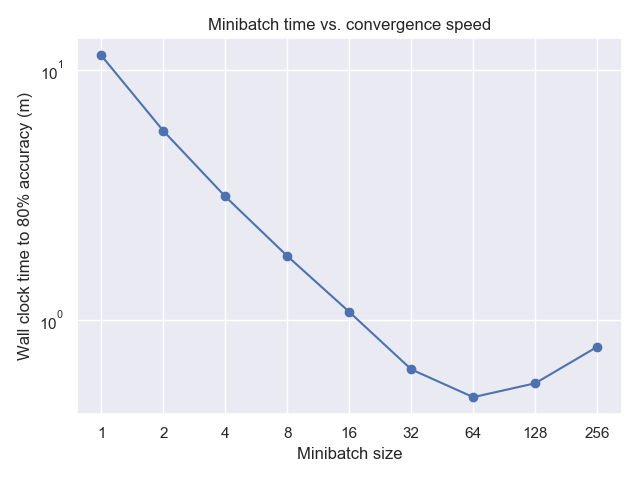}
        \caption{Wall-clock time comparison}
        \label{subfig:max_training_time}
    \end{subfigure}
    \caption{(a) Accuracy curves for networks with $n_\textrm{neurons} = 100$ over the course of training for various settings of batch size, with parameter updates computed by taking the maximum of individual contributions over the minibatch dimension. Accuracy curves are smoothed with a Hann filter of length 10. Wall-clock time needed to reach 80\% test set accuracy is plotted in (b).}
\end{figure*}

Fixing $n_\textrm{neurons} = 100$ and varying the batch size in $[1, 2, \ldots, 256]$, we investigate how the originally serial method performs in the minibatched setting. Output neurons are re-labeled and accuracy on the test dataset is assessed after every 250 training examples. With larger batch sizes, the network fails to learn to classify the data with the default parameter reduction: averaging parameter updates across the minibatch dimension (data not shown). On the other hand, Figure \ref{subfig:max_training_curves} shows that this issue can be partially mitigated by utilizing a different parameter reduction method: taking the per-parameter (synapse) maximum on each time step (the ``maximum'' method). With this reduction, the networks achieve a comparable maximum test accuracy regardless of batch size.

We conjecture this mismatch in accuracy is due to there being more examples per minibatch then there are output neurons; as described above, one neuron typically ``wins'' per example in the soft WTA. Therefore, with more inputs than neurons, there must exist at least one neuron which fires for two or more different examples in the minibatch, leading to conflicting weight updates that may cancel each other out. Using the per-parameter maximum partially solves this problem by discarding smaller weights updates, allowing the larger updates to coalesce and enabling learning of coherent synapse weights. Still, there is a non-negligible loss in accuracy with moderately large batch sizes, and this problem is exacerbated with increasingly larger batch sizes (data not shown). An interesting direction of future work is to investigate training methods that are more robust to the choice of batch size.

\begin{figure}
    \centering
    \begin{subfigure}[t]{0.225\textwidth}
        \centering
        \includegraphics[width=\textwidth]{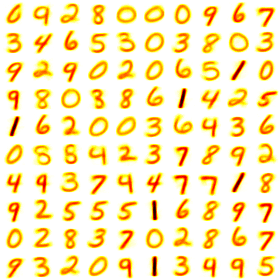}
        \caption{Batch size = 1}
        \label{subfig:weights_max_1}
    \end{subfigure}
    ~
    \begin{subfigure}[t]{0.225\textwidth}
        \centering
        \includegraphics[width=\textwidth]{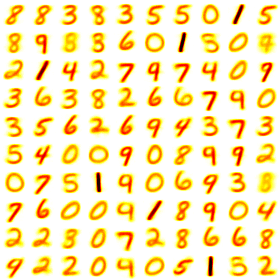}
        \caption{Batch size = 256}
        \label{subfig:weights_max_256}
    \end{subfigure}
    \caption{Filter weights learned by networks trained with the ``maximum'' reduction method and (a) serial updates (batch size 1) and (b) a large degree of data parallelism (batch size 256).}
    \label{fig:weights_comparison}
\end{figure}

Figure \ref{subfig:max_training_time} compares the wall-clock time required to reach 80\% accuracy accuracy with the same network trained with various batch sizes. In particular, with batch size 1, nearly 12 minutes is required to reach this accuracy level. With batch size 64, less than 30 seconds are needed, a $\approx$24X speedup, with only a small loss in maximum performance. Figure \ref{fig:weights_comparison} compares the weights learned with different settings of the batch size. Importantly, visual inspection reveals very few qualitative differences between the learned filters. This suggests that, with proper tuning of the hyper-parameters of the classification part of the method, networks trained with larger minibatch sizes may attain classification performance equal to that of the serial method.

\section{Discussion}
\label{sec:discussion}

Our implementation can be extended to arbitrarily complex neuron and synapse models. The user may subclass \texttt{BindsNET}'s \texttt{Nodes} or \texttt{Connection} objects, and then specify which neuronal variables need to be duplicated across the batch dimension. As discussed before, per-synapse variables typically require more memory than per-neuron variables, and each batched variable will require its own memory resources. For this reason, networks with simplistic neuron and synapse types can often be enlarged and parallelized to a greater degree than networks with more complex components.

Although our focus in the exposition and experiments of this paper has been on GPU-based simulation, minibatch processing can also be used with CPUs. However, the reductions in wall-clock time from using this approach are much less drastic than simulating with GPUs.

For any given task, the careful selection of a reduction of parameter updates across the minibatch dimension may be needed to achieve the desired learning outcome. In online learning rules, since synapse weights are triggered in an event-based fashion, these may be sparse in time, so taking the average update among many zeros may result in slow learning. For that reason, users are free to select or implement reduction methods that suit their particular learning problem.

\section{Conclusion}
\label{sec:conclusion}

Spiking neural networks are rapidly becoming viable tools for investigations into powerful, biologically plausible forms of machine learning \cite{Pfeiffer2018}. While processing batches of data in parallel in real neural circuits may not be plausible, in simulation, it serves as a useful optimization for the sake of computational efficiency. To date, the bulk of simulation has been implemented as serial processes, which often does not scale to large datasets: the speed of research iteration is extremely low due to the high cost of running even a single pass through the data. Thus, we have introduced and demonstrated the utility of a general-purpose implementation of minibatch processing in SNNs that can be leveraged to reduce simulation run-times and increase the speed of iteration of research ideas. With enough GPU memory and the proper choice of minibatch size, the wall-clock time of any simulation can be significantly reduced while preserving learning capabilities; we believe this is an important technological milestone in the effort to leverage spiking neural networks in modeling studies and machine learning experiments alike.

\section{Acknowledgements}

We would like to thank Sam Wenke, Jim Fleming, and Mike Qiu for their careful review of the manuscript.

\bibliographystyle{aaai}
\bibliography{main}




\end{document}